\documentclass[conference]{IEEEtran}
\usepackage{cite}
\usepackage{amsmath,amssymb,amsfonts}
\usepackage{graphicx}
\usepackage{booktabs,tabularx,array}
\usepackage{url}
\usepackage[hidelinks]{hyperref}
\hypersetup{pdftitle={Shared KV Heads, Sliding Windows, and Vocabulary Size: Measured Memory and Retrieval Trade-offs in LLaMA 7B, Mistral 7B, and Qwen2.5-7B}, pdfauthor={Kapardhi Kannekanti; Durvankur Chavan; B Tarun Sandilya}}
\usepackage{textcomp}
\newcolumntype{Y}{>{\raggedright\arraybackslash}X}
\begin{document}

\title{Shared KV Heads, Sliding Windows, and Vocabulary Size:\\Measured Memory and Retrieval Trade-offs in\\LLaMA 7B, Mistral 7B, and Qwen2.5-7B}

\author{
\IEEEauthorblockN{Kapardhi Kannekanti}
\IEEEauthorblockA{\texttt{kapardhikannekanti@gmail.com}}
\and
\IEEEauthorblockN{Durvankur Chavan}
\IEEEauthorblockA{\texttt{durvankurchavan4@gmail.com}}
\and
\IEEEauthorblockN{B Tarun Sandilya}
\IEEEauthorblockA{\texttt{balivadatarun@gmail.com}}
}
\maketitle
\begin{abstract}
Three widely used 7B-class language models, original LLaMA 7B, Mistral 7B v0.1, and Qwen2.5-7B, differ in three design choices that decide what fits on a small GPU: how many key-value (KV) heads the attention layers keep (32, 8, and 4), whether attention is limited to a sliding window (4,096 tokens in Mistral), and how large the vocabulary is (32,000 versus 151,643 tokens). We measure the consequences of each choice on one 16\,GB T4 GPU. (1) Passkey retrieval in Mistral stops at the window. In 134 trials with the key placed 3,600 to 4,600 tokens from the end of the prompt, Mistral retrieved it in all 72 trials where it lay within 4,089 tokens and in none of the 62 trials where it lay 4,098 or more tokens away. With the same weights and the window turned off, retrieval succeeds at every tested position in 8K-token prompts, so the mask causes the limit; at 16K and 32K tokens the window-free model fails completely, while the windowed model still retrieves nearby keys. Qwen2.5 retrieves the key in all 60 trials up to 32K tokens. (2) KV-cache memory per token is 512, 128, and 56\,KiB. LLaMA runs out of memory at 32K tokens, Qwen2.5 holds 64K tokens in 3.5\,GiB, and Mistral's cache stops at 512\,MiB. (3) On 1,997 parallel sentences in 15 languages, Qwen2.5's tokenizer removes 38--58\% of tokens for five languages and 3--11\% for the other ten; combined with KV sharing, LLaMA needs 9.5--22.0 times more cache memory than Qwen2.5 for the same sentence. (4) Decode latency in a common runtime grows with cache size, and copying the cache, measured on its own, costs about as much as that growth. Code and results are public.
\end{abstract}


\section{Introduction}
Open language models of about seven billion parameters run comfortably on a single commodity GPU with 16\,GB of memory. On such hardware, three design choices decide how much context a model can hold and use: the number of key-value (KV) heads, which sets the size of the KV cache; the attention mask, which sets how far back a layer can look; and the vocabulary, which sets how many tokens a text needs.

We study three models that differ cleanly on these three choices. Original LLaMA 7B~\cite{llama} uses one KV head per query head, full causal attention, and a 32,000-token vocabulary. Mistral 7B v0.1~\cite{mistral} keeps the same width and depth but shares each KV head among four query heads and limits attention to a 4,096-token sliding window. Qwen2.5-7B~\cite{qwen} shares each KV head among seven query heads, uses full attention with long-context training, and has a vocabulary of 151,643 regular tokens. All three are base checkpoints, and all three fit on one T4 GPU with 4-bit weights. Newer models keep these mechanisms in different combinations: Llama 3 adopted grouped KV heads, and Mistral removed the sliding window in v0.2~\cite{mistralv02}. The three models therefore serve as a compact set of cases for design choices that remain in use.

This paper reports measurements. Its findings are:
\begin{itemize}
\item \textbf{The sliding window sets a retrieval limit at the window boundary, and the mask is its cause.} Mistral retrieves a passkey in 72 of 72 trials when the key lies within 4,089 tokens of the end of the prompt and in 0 of 62 trials when it lies 4,098 or more tokens away. With the window turned off at inference, on the same weights, retrieval succeeds at every tested position in 8K-token prompts.
\item \textbf{The window also keeps the model usable at long lengths.} Without the window, Mistral fails every trial at 16K and 32K tokens, including keys that are close to the question. With the window, it still retrieves every key inside the window at those lengths.
\item \textbf{KV sharing and vocabulary compound.} Per token, LLaMA's cache is 9.1 times larger than Qwen2.5's. Per sentence, it is 9.5 to 22.0 times larger, because LLaMA also needs more tokens for the same text.
\item \textbf{Latency growth in a common runtime is partly an implementation cost.} Copying the cache at each step, measured alone, costs about 17--20 ms per GiB of cache, which is comparable to the whole latency growth for LLaMA.
\end{itemize}
We make no claim about which design gives better benchmark quality. The three models differ in training data and training length, and no measurement here separates those factors.

\section{Background and Related Work}
\subsection{Shared Structure of the Three Models}
All three models are causal decoder-only Transformers~\cite{transformer} with pre-normalization by RMSNorm~\cite{rmsnorm}, rotary position embeddings (RoPE)~\cite{rope}, and SwiGLU feed-forward networks (FFNs)~\cite{swiglu}. Table~\ref{tab:arch} lists the dimensions, taken from the papers and the published checkpoint configurations~\cite{qwenconfig,mistralconfig,llamaconfig}. For width $d$, FFN width $d_{\mathrm{ff}}$, and $L$ blocks, the FFN weights number about
\begin{equation}
N_{\mathrm{FFN}}=3\,d\,d_{\mathrm{ff}}\,L,
\label{eq:ffn}
\end{equation}
which gives 4.329B, 5.637B, and 5.703B for LLaMA, Mistral, and Qwen2.5. Qwen2.5 has a wider FFN per block but a narrower stream and fewer blocks, so its stack-wide FFN count is only 1.17\% above Mistral's.

\begin{table*}[!t]
\caption{Architecture of the Three Checkpoints, From the Source Papers and Published Configurations~\cite{llama,mistral,qwen,qwenconfig,mistralconfig,llamaconfig}.}
\label{tab:arch}
\centering
\footnotesize
\renewcommand{\arraystretch}{1.1}
\begin{tabularx}{\textwidth}{@{}>{\raggedright\arraybackslash}p{0.17\textwidth}YYY@{}}
\toprule
\textbf{Property} & \textbf{Original LLaMA 7B} & \textbf{Mistral 7B v0.1} & \textbf{Qwen2.5-7B base} \\
\midrule
Tokenizer / embedding slots & SentencePiece BPE with byte fallback; 32,000 slots & SentencePiece-family BPE with byte fallback; 32,000 slots & Byte-level BPE; 151,643 regular tokens; 152,064 embedding slots \\
Residual width $d$ / blocks $L$ & 4,096 / 32 & 4,096 / 32 & 3,584 / 28 \\
Attention & Full causal & Sliding window, $W=4{,}096$ & Full causal \\
Query / KV heads & 32 / 32 & 32 / 8 & 28 / 4 \\
Head dimension & 128 & 128 & 128 \\
FFN width $d_{\mathrm{ff}}$ (SwiGLU) & 11,008 & 14,336 & 18,944 \\
Input--output weight tying & Untied & Untied & Untied \\
Sequence length & 2,048-token configuration & 8,192 listed in paper & 32,768 in late pretraining; 131,072 supported with extension methods \\
\bottomrule
\end{tabularx}
\end{table*}

\subsection{KV Sharing}
\begin{figure*}[!t]
\centering
\includegraphics[width=0.82\textwidth]{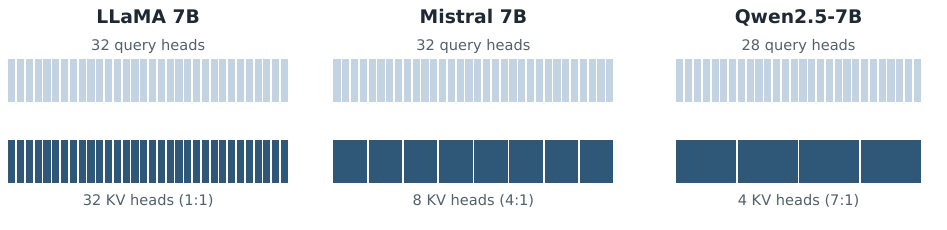}
\caption{Query--KV head sharing in original LLaMA 7B, Mistral 7B, and Qwen2.5-7B, drawn to scale. Each KV head (dark) is shared by the query heads (light) directly above it.}
\label{fig:heads}
\end{figure*}

Multi-head attention keeps one key and one value projection per query head. Multi-query attention shares a single KV head among all query heads~\cite{mqa}, and grouped-query attention (GQA) shares each KV head among a group~\cite{gqa}. Fig.~\ref{fig:heads} shows the three configurations. During generation, the keys and values of earlier tokens are cached. For $T$ cached positions the cache occupies
\begin{equation}
M_{\mathrm{KV}}=2\,L\,H_{kv}\,d_h\,b\,T
\label{eq:kv}
\end{equation}
bytes, where $H_{kv}$ is the number of KV heads, $d_h$ the head dimension, and $b$ the bytes per element. At $b=2$ (FP16), one cached token costs 512\,KiB in LLaMA, 128\,KiB in Mistral, and 56\,KiB in Qwen2.5.

\subsection{Sliding-Window Attention}
\begin{figure}[!t]
\centering
\includegraphics[width=\columnwidth]{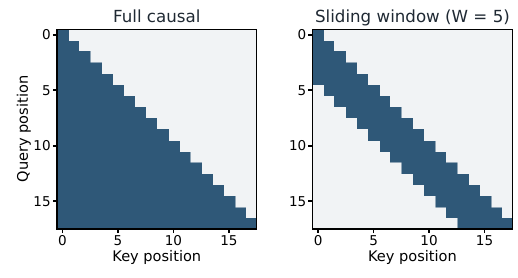}
\caption{Full causal and sliding-window masks (illustrative $W=5$). Dark cells are directly accessible query--key pairs.}
\label{fig:mask}
\end{figure}

Local attention limits each position to the previous $W$ positions (Fig.~\ref{fig:mask}). Transformer-XL~\cite{transformerxl} and Longformer~\cite{longformer} used related ideas to extend context at bounded cost. Mistral 7B v0.1 uses $W=4{,}096$ and lists a context length of 8,192~\cite{mistral}. Because no layer attends beyond $W$ positions, the cache can be a rolling buffer of $W$ entries, and $T$ in~\eqref{eq:kv} is capped at $W$. The Mistral authors note that stacked layers can pass information beyond one window, which gives a theoretical attention span of about 131K tokens at the last layer~\cite{mistral}. Mistral 7B v0.2 removed sliding-window attention and raised the context length to 32K tokens~\cite{mistralv02}. Assogba and Ren~\cite{assogba} found that Mistral-7B-v0.1 fails to retrieve a key function more than about 4K tokens from the generation site in 8K-token code prompts, and confirmed the link by changing the window size at inference. Wang et al.~\cite{wange5} report the same 4K limit on passkey retrieval for an embedding model fine-tuned from Mistral, and worse results when the window is widened to 32K tokens. Section IV-C extends these results: it locates the limit between 4,089 and 4,098 tokens, turns the window off on the same weights at 8K to 32K tokens, and repeats the test at two weight precisions.

\subsection{Long-Context Evaluation}
Passkey retrieval hides one short fact in filler text and asks for it~\cite{landmark}. Broader tests show that the usable context of a model is often shorter than its advertised context and depends on where the relevant text sits~\cite{lostmiddle,ruler,longbench}. Qwen2.5 reports support for 131,072 tokens after training at up to 32,768 tokens and applying YaRN~\cite{yarn} and Dual Chunk Attention~\cite{dca}; we test the base checkpoint up to 32K tokens with no RoPE scaling enabled: the base configuration sets \texttt{rope\_scaling} to null and the RoPE base to 1,000,000~\cite{qwen,qwenconfig}.

\subsection{Tokenization Cost}
LLaMA and Mistral use SentencePiece-based BPE vocabularies of 32,000 tokens with byte fallback~\cite{bpe,sentencepiece,llama,mistraltoken}; Qwen2.5 uses a byte-level BPE vocabulary inherited from earlier Qwen models~\cite{qwen}. Prior work shows that tokenizers encode some languages with far more tokens than others~\cite{petrov,ahia,rust}. We apply the same parallel-text method to these three tokenizers and connect it to cache memory through~\eqref{eq:kv}. A larger vocabulary also costs parameters: with untied input and output matrices, Qwen2.5 spends about 545 million weights on each, against 131 million for LLaMA and Mistral.

\section{Experimental Setup}
\label{sec:setup}
\textbf{Models.} We used the base checkpoints \texttt{mistralai/Mistral-7B-v0.1}, \texttt{Qwen/Qwen2.5-7B}, and \texttt{huggyllama/llama-7b} from the Hugging Face Hub. The LLaMA weights are a community mirror of the original release. The repository revisions and the SHA-256 hashes of the weight files are recorded in our result files.

\textbf{Hardware and software.} All GPU runs used one NVIDIA T4 with 14.56\,GiB of usable memory (free Google Colab tier), transformers 5.22.2~\cite{transformers}, PyTorch 2.14.1~\cite{pytorch} with CUDA 13.0, and bitsandbytes 0.50.2. Linear-layer weights were quantized to 4-bit NF4~\cite{qlora} with FP16 compute; embeddings, output head, and KV cache stay in FP16. Attention used PyTorch's scaled-dot-product kernels and the library's dynamic cache. Prompts were prefilled in chunks of 1,024 tokens, and decoding was greedy at batch size one.

\textbf{Sliding window in this runtime.} For Mistral, this library version applies the sliding-window mask in every layer and keeps the most recent $W-1$ positions in the cache. We checked on a small random model that chunked prefill gives the same logits as one full pass, with and without the window.

\textbf{Passkey prompts.} A five-digit key is stated in the sentence ``The pass key is $k$. Remember it. $k$ is the pass key.''\ inside repeated filler sentences, and the prompt ends with ``What is the pass key? The pass key is''~\cite{landmark}. A trial succeeds if the first number in the greedy continuation equals the key. Each trial uses a different key and a different starting point in the filler cycle; for each setting, every model sees the same keys. Lengths are counted in each model's own tokens. We report the distance from the nearer of the two key mentions to the end of the prompt; it is exact to within one token.

\textbf{Tokenization.} We encode the 1,997 sentences of the NTREX-128 news test set~\cite{ntrex} in 15 languages with each model's tokenizer, without special tokens. NTREX-128 contains professional translations of the same English source, so counts are comparable across languages.

\section{Results}
\subsection{Tokenization Cost}
\label{sec:tok}
Table~\ref{tab:tok} reports tokens per sentence. Bootstrap 95\% intervals for the change column, resampling sentences, are within $\pm0.4$ percentage points for every language.

\begin{table}[!t]
\caption{Tokens per Sentence on 1,997 Parallel NTREX-128 Sentences~\cite{ntrex}. $\Delta$: Change (\%) for Qwen2.5 Relative to the Better of LLaMA and Mistral. BF: Byte-Fallback Share (\%) for LLaMA/Mistral.}
\label{tab:tok}
\centering
\footnotesize
\setlength{\tabcolsep}{4pt}
\begin{tabular}{@{}lrrrrr@{}}
\toprule
\textbf{Language} & \textbf{LLaMA} & \textbf{Mistral} & \textbf{Qwen2.5} & $\Delta$ & \textbf{BF} \\
\midrule
English & 30.5 & 29.3 & 26.6 & $-9.3$ & 0/0 \\
German & 42.6 & 45.6 & 41.0 & $-3.8$ & 0/0 \\
French & 44.2 & 45.9 & 41.4 & $-6.3$ & 0/0 \\
Spanish & 42.0 & 43.6 & 38.4 & $-8.5$ & 0/0 \\
Russian & 50.5 & 55.2 & 48.2 & $-4.6$ & 0/0 \\
Vietnamese & 87.0 & 83.8 & 38.1 & $-54.6$ & 7/1 \\
Arabic & 106.2 & 104.4 & 44.2 & $-57.6$ & 0/0 \\
Chinese & 61.4 & 48.1 & 27.2 & $-43.3$ & 48/20 \\
Japanese & 67.4 & 62.3 & 38.4 & $-38.3$ & 25/15 \\
Korean & 98.2 & 73.7 & 45.7 & $-37.9$ & 51/18 \\
Hindi & 141.9 & 135.7 & 121.2 & $-10.7$ & 12/5 \\
Bengali & 157.3 & 138.3 & 130.7 & $-5.5$ & 33/17 \\
Telugu & 311.8 & 203.0 & 182.4 & $-10.2$ & 90/58 \\
Tamil & 184.4 & 176.4 & 170.4 & $-3.4$ & 24/18 \\
Swahili & 56.1 & 56.4 & 51.5 & $-8.2$ & 0/0 \\
\bottomrule
\end{tabular}
\end{table}

Qwen2.5's tokenizer and vocabulary design produce uneven reductions across languages. Relative to the better 32,000-token vocabulary, it removes 38--58\% of tokens for Vietnamese, Arabic, Chinese, Japanese, and Korean, but only 3--11\% for the other ten languages. For the four Indic languages, all three tokenizers produce about one token per character or more: Qwen2.5 encodes 0.7 to 1.1 characters per token there, against 4.7 for English.

Allocation matters at equal size. LLaMA and Mistral both have 32,000 slots, yet Mistral uses 34.9\% fewer tokens than LLaMA for Telugu, 25.0\% fewer for Korean, and 21.7\% fewer for Chinese, while using 3.9--9.3\% more for German, French, Spanish, and Russian. For Telugu, 90\% of LLaMA's tokens are byte-fallback tokens, against 58\% for Mistral.

Vocabulary and KV sharing compound. Multiplying the measured token counts by the per-token cache size from~\eqref{eq:kv} shows that LLaMA needs 9.5 to 22.0 times more cache memory than Qwen2.5 to hold the same sentence, with the largest ratios for Arabic (22.0), Vietnamese (20.9), Chinese (20.6), and Korean (19.6). This ratio is a calculation from measured counts, not a separate GPU measurement.

\subsection{KV-Cache Memory}
\label{sec:mem}
\begin{figure*}[!t]
\centering
\includegraphics[width=\textwidth]{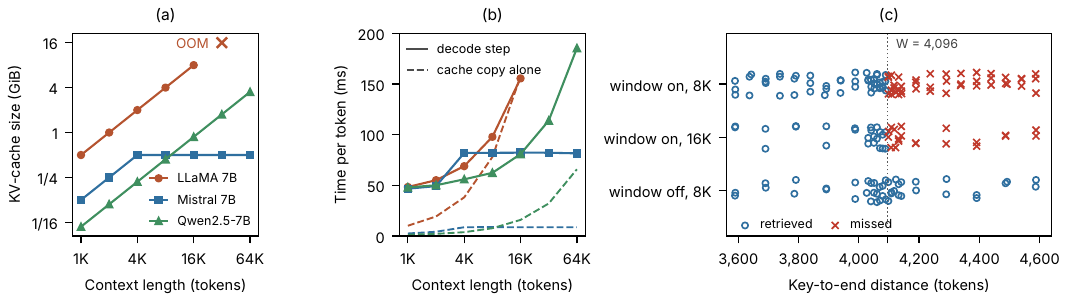}
\caption{Measurements on one T4 GPU with 4-bit weights. (a) KV-cache size; LLaMA runs out of memory at 32K tokens. (b) Median decode time per token (solid) and the time of the per-step cache concatenation measured alone (dashed). (c) Mistral passkey trials by the distance from the key to the end of the prompt, with vertical jitter. With the window on, every miss lies beyond 4,096 tokens; with the window off, every trial succeeds.}
\label{fig:meas}
\end{figure*}

We prefilled random token sequences of 1K to 64K tokens and recorded the bytes held by the cached tensors and the peak device memory above the model weights (Fig.~\ref{fig:meas}(a), Table~\ref{tab:gpu}). As a check, the cached tensors have exactly the size that~\eqref{eq:kv} gives. Mistral's cache stops growing at 4,095 positions (511.9\,MiB) and stays there through 64K tokens. LLaMA's cache reaches 8\,GiB at 16K tokens; at 32K tokens, where the cache alone would need 16\,GiB, the run fails with an out-of-memory error. Qwen2.5 holds 64K tokens in a 3.5\,GiB cache.

Peak memory is what decides whether a prompt fits, and~\eqref{eq:kv} does not predict it. For Qwen2.5 at 64K tokens the peak is 4.60\,GiB above the weights, 31\% more than the cache. Prefill throughput follows the attention pattern: from 1K to 64K tokens Mistral drops from 1,081 to 793 tokens/s, while Qwen2.5, with full attention, drops from 1,202 to 394 tokens/s.

\begin{table*}[!t]
\caption{GPU Measurements on One T4 (4-Bit NF4 Weights, FP16 KV Cache, Batch Size One). Peak: Device Memory Above the Model Weights. Decode: Median of Five Blocks of 32 Steps. Cache Copy: Per-Step Cache Concatenation Measured Alone. OOM: Out of Memory; --: Not Run. L: LLaMA 7B; M: Mistral 7B; Q: Qwen2.5-7B.}
\label{tab:gpu}
\centering
\footnotesize
\setlength{\tabcolsep}{5pt}
\begin{tabular}{@{}lrrrrrrrrr@{}}
\toprule
 & \multicolumn{3}{c}{\textbf{Peak memory (GiB)}} & \multicolumn{3}{c}{\textbf{Decode (ms/token)}} & \multicolumn{3}{c}{\textbf{Cache copy (ms)}} \\
\cmidrule(lr){2-4}\cmidrule(lr){5-7}\cmidrule(l){8-10}
\textbf{Context} & L & M & Q & L & M & Q & L & M & Q \\
\midrule
1K & 0.66 & 0.32 & 0.28 & 48.4 & 46.7 & 47.9 & 10.1 & 2.4 & 1.2 \\
2K & 1.16 & 0.45 & 0.35 & 55.1 & 49.2 & 50.1 & 19.3 & 4.2 & 2.1 \\
4K & 2.16 & 0.70 & 0.45 & 68.9 & 82.0 & 55.9 & 38.1 & 8.6 & 3.7 \\
8K & 4.17 & 0.83 & 0.68 & 97.9 & 82.1 & 62.3 & 78.0 & 8.7 & 7.6 \\
16K & 8.18 & 0.83 & 1.18 & 155.6 & 82.3 & 80.9 & 157.6 & 8.6 & 15.9 \\
32K & OOM & 0.83 & 2.32 & -- & 82.2 & 113.9 & -- & 8.6 & 32.1 \\
64K & -- & 0.83 & 4.60 & -- & 81.6 & 185.4 & -- & 8.6 & 65.8 \\
\bottomrule
\end{tabular}
\end{table*}

\subsection{Passkey Retrieval}
\label{sec:passkey}
\textbf{Standard grid.} Table~\ref{tab:passkey} gives the results for keys at 10, 30, 50, 70, and 90\% depth, three keys per depth. Qwen2.5 retrieved the key in all 60 trials, including keys 29,450 tokens from the end of the prompt. LLaMA succeeded in all 30 trials at 1K and 2K tokens and in none at 4K tokens, beyond its 2,048-token configuration, where it produced strings of periods. Mistral succeeded in all 15 trials at 4K tokens, in 8 at 8K, and in 3 each at 16K and 32K. Its outcome depends on the position of the key, not on the prompt length: across all lengths, it retrieved 29 of the 30 keys that lay within 4,096 tokens of the end and none of the 30 that lay further away. The one miss inside the window dropped a digit (``9470'' for 94970); the same trial succeeded with 8-bit weights.

\begin{table}[!t]
\caption{Passkey Retrieval: Keys Retrieved / Trials. Five Depths per Length; Three Keys per Depth with 4-Bit Weights, Two with 8-Bit Weights. --: Not Run.}
\label{tab:passkey}
\centering
\footnotesize
\setlength{\tabcolsep}{3.5pt}
\begin{tabular}{@{}lcccccc@{}}
\toprule
 & \textbf{LLaMA} & \multicolumn{3}{c}{\textbf{Mistral}} & \multicolumn{2}{c}{\textbf{Qwen2.5}} \\
\cmidrule(lr){3-5}\cmidrule(l){6-7}
\textbf{Length} & 4-bit & 4-bit & window off & 8-bit & 4-bit & 8-bit \\
\midrule
1K & 15/15 & -- & -- & -- & -- & -- \\
2K & 15/15 & -- & -- & -- & -- & -- \\
4K & 0/15 & 15/15 & 15/15 & -- & 15/15 & -- \\
8K & -- & 8/15 & 15/15 & 6/10 & 15/15 & -- \\
16K & -- & 3/15 & 0/15 & 2/10 & 15/15 & 10/10 \\
32K & -- & 3/15 & 0/15 & -- & 15/15 & 10/10 \\
\bottomrule
\end{tabular}
\end{table}

\textbf{Distance sweep.} To locate the limit, we placed the key 3,600 to 4,600 tokens from the end of the prompt, in steps of about 10 tokens near 4,096, in prompts of 8K tokens (90 trials) and 16K tokens (44 trials). Fig.~\ref{fig:meas}(c) shows the result. Mistral retrieved the key in all 72 trials in which the nearer mention lay within 4,089 tokens of the end and in none of the 62 trials in which it lay 4,098 or more tokens away. The window of 4,096 tokens falls between these two distances, and the result is the same at both prompt lengths. In five trials only the second mention of the key was inside the window; the key was still correct in all five, although the text after it no longer repeated ``Remember it''. Just outside the window, the model usually answers with a ten-digit sequence such as ``1234567890''; far  outside it, the model answers ``the password'' or ``the pass key''.

\textbf{Window off.} To test whether the mask causes the limit, we loaded the same weights with the sliding window disabled, which gives full causal attention and a full cache. At 8K tokens, the listed context length of the model, retrieval succeeded in all 15 grid trials and in all 44 sweep trials, including every distance at which the windowed model failed. The limit at 4,096 tokens is therefore an effect of the attention mask at inference, not of a limit in what the weights can retrieve inside 8K tokens.

Beyond 8K tokens the result reverses. Without the window, Mistral failed all 30 trials at 16K and 32K tokens and produced degenerate text (repetitions of ``Question''), even for keys about 1,650 tokens from the end. With the window, the model retrieved all six keys inside the window at these lengths. In this model, the window removes retrieval beyond 4,096 tokens, and it also keeps every layer inside the range of relative positions seen in training, which is consistent with the window keeping the model functional on prompts longer than its training context, although our experiments do not isolate the positional mechanism. The stacked-window span of about 131K tokens~\cite{mistral} did not yield retrieval in any of the 92 trials with the key outside the window.

\textbf{Weight precision.} With 8-bit weights, Mistral retrieved 8 of the 8 keys inside the window and 0 of the 12 outside it, and Qwen2.5 retrieved 20 of 20. The retrieval pattern does not depend on 4-bit quantization.

\subsection{Decode Latency}
At 1K tokens, one decode step takes 47--48\,ms for all three models (Table~\ref{tab:gpu}, Fig.~\ref{fig:meas}(b)). At 16K tokens it takes 155.6\,ms for LLaMA, 82.3\,ms for Mistral, and 80.9\,ms for Qwen2.5; at 64K tokens Qwen2.5 needs 185.4\,ms and Mistral 81.6\,ms. Repeated blocks agree closely: from 4K tokens upward, the five block medians at each point differ by less than 4\,ms.

Part of this growth is a property of the runtime. The library's dynamic cache concatenates each layer's key and value tensors with the new position at every step, which copies the whole cache. Measured alone, this copy takes about 17--20 ms per GiB of cache: 157.6\,ms for LLaMA at 16K tokens, 65.8\,ms for Qwen2.5 at 64K tokens, and 8.6\,ms for Mistral's full buffer. For LLaMA the copy time is as large as the whole growth in step time, and at 16K tokens the isolated measurement slightly exceeds the full step, so it is an upper estimate of the cost inside a step. For Qwen2.5 it accounts for about half of the growth from 1K to 64K tokens. A runtime with a preallocated cache would not pay this cost, so the latency ordering of the three designs in such a runtime may differ from the one in Table~\ref{tab:gpu}.

Mistral's step time rises from 49.2\,ms at 2K to 82.0\,ms at 4K tokens, when its buffer first fills, and stays flat afterward. Cache copying explains about 4\,ms of this 33\,ms step. We did not isolate the rest, which occurs when the sliding-window path of the runtime becomes active.

\section{Discussion}
For a deployment on a 16\,GB GPU, the three designs give three different limits. LLaMA's unshared KV heads make memory the limit: 16K tokens use more than 8\,GiB, and its 2,048-token configuration makes longer prompts unusable in any case. Mistral's window makes memory and step time constant beyond 4K tokens, and the model stays functional on prompts of 32K tokens, but a fact that is more than 4,096 tokens back was not retrieved in any of our trials. Qwen2.5's 7:1 sharing lets full attention reach 64K tokens in 3.5\,GiB of cache, and retrieval held at every tested distance; its costs are a peak memory that exceeds the cache, a step time that grows with length, and about 830 million more embedding and output weights than the 32,000-token models.

The window-off experiment separates two effects that the windowed model combines. Inside the training length, the mask alone limits retrieval. Beyond the training length, removing the mask led to degenerate output at 16K and 32K tokens in our trials. This is consistent with the later change in the same model family: Mistral 7B v0.2 removed the window and was trained for a 32K context~\cite{mistralv02}. Our measurements suggest that removing the window from v0.1 at inference is useful only up to 8K tokens.

\section{Limitations}
The three checkpoints differ in training data, training length, and positional settings, so differences between models cannot be attributed to one design choice. Only the window-off experiment is an intervention, and it changes inference on a model that was trained with the window.

Passkey retrieval is a synthetic test with repetitive filler. It shows whether a model can recover one exact string, not how well it uses long documents. We tested Qwen2.5 only up to 32K tokens. Distances are exact to within one token, and no trial placed the nearer key mention between 4,090 and 4,097 tokens, so the limit is located to within that interval.

All GPU measurements use one T4 GPU, batch size one, 4-bit weights, and the attention and cache code of one library version. Absolute latency depends on these choices. The cache-copy time was measured outside the decode step and can overestimate the cost inside it. Peak memory and prefill throughput were measured once per point. The cause of Mistral's latency step at 4K tokens was not isolated.

The tokenization measurement covers one domain, news, and text translated from English~\cite{ntrex}. Token counts measure sequence length, not model quality. The LLaMA weights come from a community mirror; we record their hashes but did not compare them with the original release.

\section{Conclusion}
On one 16\,GB GPU, three 7B models show how KV sharing, the attention mask, and vocabulary size turn into concrete limits. KV sharing sets cache memory as the formula predicts, from 512 to 56\,KiB per token, and differences in token counts across the tokenizers widen that gap to as much as 22 times per sentence. Mistral's sliding window caps memory and step time, sets a retrieval limit that we located between 4,089 and 4,098 tokens, and is the cause of that limit inside the model's 8K context; beyond that context, removing the window made the model fail in our trials. Qwen2.5 retrieved the key at every tested distance up to 32K tokens. In a widely used runtime, copying the cache, measured on its own, costs about as much as the latency growth with context, which points to the implementation rather than the architectures.

\section*{Code and Data Availability}
The experiment code, the result files with the generated text of every trial, and the figure script are available at \url{https://github.com/professionaltarun2004/Block-Level-Comparison-of-7B-models}. NTREX-128 is available from its authors~\cite{ntrex}.

\section*{Use of Generative AI Tools}
The authors used Claude (Anthropic), a generative AI assistant, to help draft, edit, and revise the manuscript text, write the code for the measurements and figures, and check references against their published records. The authors ran all experiments, reviewed all text and code, verified every reported number against the raw measurement outputs and every reference against its source, and take full responsibility for the content of this paper.


\begin{thebibliography}{00}
\bibitem{llama} H. Touvron \emph{et al.}, ``LLaMA: Open and Efficient Foundation Language Models,'' arXiv:2302.13971v1, 2023. [Online]. Available: \url{https://arxiv.org/abs/2302.13971}
\bibitem{mistral} A. Q. Jiang \emph{et al.}, ``Mistral 7B,'' arXiv:2310.06825v1, 2023. [Online]. Available: \url{https://arxiv.org/abs/2310.06825}
\bibitem{qwen} Qwen Team, ``Qwen2.5 Technical Report,'' arXiv:2412.15115v2, 2025. [Online]. Available: \url{https://arxiv.org/abs/2412.15115}
\bibitem{mistralv02} Mistral AI, ``Mistral-7B-Instruct-v0.2,'' model card, Hugging Face, accessed Nov. 16, 2026. [Online]. Available: \url{https://huggingface.co/mistralai/Mistral-7B-Instruct-v0.2/tree/63a8b081895390a26e140280378bc85ec8bce07a}
\bibitem{transformer} A. Vaswani \emph{et al.}, ``Attention Is All You Need,'' in \emph{Advances in Neural Information Processing Systems}, 2017.
\bibitem{rmsnorm} B. Zhang and R. Sennrich, ``Root Mean Square Layer Normalization,'' in \emph{Advances in Neural Information Processing Systems}, 2019.
\bibitem{rope} J. Su \emph{et al.}, ``RoFormer: Enhanced Transformer with Rotary Position Embedding,'' arXiv:2104.09864, 2021.
\bibitem{swiglu} N. Shazeer, ``GLU Variants Improve Transformer,'' arXiv:2002.05202, 2020.
\bibitem{qwenconfig} Qwen Team, ``Qwen2.5-7B model configuration,'' Hugging Face model repository, accessed Nov. 6, 2026. [Online]. Available: \url{https://huggingface.co/Qwen/Qwen2.5-7B/blob/d149729398750b98c0af14eb82c78cfe92750796/config.json}
\bibitem{mistralconfig} Mistral AI, ``Mistral-7B-v0.1 model configuration,'' Hugging Face model repository, accessed Nov. 6, 2026. [Online]. Available: \url{https://huggingface.co/mistralai/Mistral-7B-v0.1/blob/27d67f1b5f57dc0953326b2601d68371d40ea8da/config.json}
\bibitem{llamaconfig} ``LLaMA 7B configuration,'' public Hugging Face checkpoint mirror, accessed Nov. 6, 2026. Community mirror of the original weights. [Online]. Available: \url{https://huggingface.co/huggyllama/llama-7b/blob/4782ad278652c7c71b72204d462d6d01eaaf7549/config.json}
\bibitem{mqa} N. Shazeer, ``Fast Transformer Decoding: One Write-Head is All You Need,'' arXiv:1911.02150, 2019.
\bibitem{gqa} J. Ainslie \emph{et al.}, ``GQA: Training Generalized Multi-Query Transformer Models from Multi-Head Checkpoints,'' in \emph{Proc. EMNLP}, 2023, pp. 4895--4901, doi: 10.18653/v1/2023.emnlp-main.298.
\bibitem{transformerxl} Z. Dai \emph{et al.}, ``Transformer-XL: Attentive Language Models Beyond a Fixed-Length Context,'' arXiv:1901.02860, 2019.
\bibitem{longformer} I. Beltagy, M. E. Peters, and A. Cohan, ``Longformer: The Long-Document Transformer,'' arXiv:2004.05150, 2020.
\bibitem{landmark} A. Mohtashami and M. Jaggi, ``Landmark Attention: Random-Access Infinite Context Length for Transformers,'' arXiv:2305.16300, 2023.
\bibitem{lostmiddle} N. F. Liu \emph{et al.}, ``Lost in the Middle: How Language Models Use Long Contexts,'' arXiv:2307.03172, 2023.
\bibitem{ruler} C.-P. Hsieh \emph{et al.}, ``RULER: What's the Real Context Size of Your Long-Context Language Models?,'' arXiv:2404.06654, 2024.
\bibitem{longbench} Y. Bai \emph{et al.}, ``LongBench: A Bilingual, Multitask Benchmark for Long Context Understanding,'' arXiv:2308.14508, 2023.
\bibitem{yarn} B. Peng \emph{et al.}, ``YaRN: Efficient Context Window Extension of Large Language Models,'' arXiv:2309.00071, 2023.
\bibitem{dca} C. An \emph{et al.}, ``Training-Free Long-Context Scaling of Large Language Models,'' in \emph{Proc. 41st Int. Conf. Machine Learning}, PMLR, vol. 235, 2024, pp. 1493--1510.
\bibitem{bpe} R. Sennrich, B. Haddow, and A. Birch, ``Neural Machine Translation of Rare Words with Subword Units,'' in \emph{Proc. ACL}, 2016.
\bibitem{sentencepiece} T. Kudo and J. Richardson, ``SentencePiece: A Simple and Language Independent Subword Tokenizer and Detokenizer for Neural Text Processing,'' in \emph{Proc. EMNLP System Demonstrations}, 2018.
\bibitem{mistraltoken} Mistral AI, ``Tokenization: Templates and Tokenizer Versions,'' official cookbook, accessed Nov. 6, 2026. [Online]. Available: \url{https://github.com/mistralai/cookbook/blob/main/concept-deep-dive/tokenization/templates.md}
\bibitem{petrov} A. Petrov, E. La Malfa, P. H. S. Torr, and A. Bibi, ``Language Model Tokenizers Introduce Unfairness Between Languages,'' in \emph{Advances in Neural Information Processing Systems}, vol. 36, 2023.
\bibitem{ahia} O. Ahia, S. Kumar, H. Gonen, J. Kasai, D. R. Mortensen, N. A. Smith, and Y. Tsvetkov, ``Do All Languages Cost the Same? Tokenization in the Era of Commercial Language Models,'' in \emph{Proc. EMNLP}, 2023, pp. 9904--9923, doi: 10.18653/v1/2023.emnlp-main.614.
\bibitem{rust} P. Rust, J. Pfeiffer, I. Vuli\'c, S. Ruder, and I. Gurevych, ``How Good is Your Tokenizer? On the Monolingual Performance of Multilingual Language Models,'' in \emph{Proc. ACL-IJCNLP}, 2021, pp. 3118--3135, doi: 10.18653/v1/2021.acl-long.243.
\bibitem{transformers} T. Wolf \emph{et al.}, ``Transformers: State-of-the-Art Natural Language Processing,'' in \emph{Proc. EMNLP: System Demonstrations}, 2020, pp. 38--45, doi: 10.18653/v1/2020.emnlp-demos.6.
\bibitem{pytorch} A. Paszke \emph{et al.}, ``PyTorch: An Imperative Style, High-Performance Deep Learning Library,'' arXiv:1912.01703, 2019.
\bibitem{qlora} T. Dettmers, A. Pagnoni, A. Holtzman, and L. Zettlemoyer, ``QLoRA: Efficient Finetuning of Quantized LLMs,'' in \emph{Advances in Neural Information Processing Systems}, vol. 36, 2023.
\bibitem{ntrex} C. Federmann, T. Kocmi, and Y. Xin, ``NTREX-128 -- News Test References for MT Evaluation of 128 Languages,'' in \emph{Proc. First Workshop on Scaling Up Multilingual Evaluation (SUMEval)}, 2022, pp. 21--24, doi: 10.18653/v1/2022.sumeval-1.4.
\bibitem{assogba} Y. Assogba and D. Ren, ``Evaluating Long Range Dependency Handling in Code Generation LLMs,'' \emph{Transactions on Machine Learning Research}, 2025.
\bibitem{wange5} L. Wang, N. Yang, X. Huang, L. Yang, R. Majumder, and F. Wei, ``Improving Text Embeddings with Large Language Models,'' in \emph{Proc. ACL}, 2024, doi: 10.18653/v1/2024.acl-long.642.
\end{thebibliography}
\end{document}